\documentclass{article} 
\usepackage{iclr2024_conference,times}

\usepackage{amsmath,amsfonts,bm}

\def\eqref#1{equation~\ref{#1}}

\def\1{\bm{1}}

\DeclareMathAlphabet{\mathsfit}{\encodingdefault}{\sfdefault}{m}{sl}
\SetMathAlphabet{\mathsfit}{bold}{\encodingdefault}{\sfdefault}{bx}{n}

\usepackage{hyperref}
\usepackage{url}
\usepackage{graphicx}

\title{Scalable Ensembling For Mitigating Reward Overoptimisation}

\author{Ahmed M. Ahmed, Rafael Rafailov, Stepan Sharkov, Xuechen Li, Sanmi Koyejo  \\
Department of Computer Science\\
Stanford University\\
\texttt{\{ahmedah, rafailov, stpshrkv, lxuechen, sanmi\}@stanford.edu} \\
}

\iclrfinalcopy 
\begin{document}

\maketitle

\begin{abstract}
Reinforcement Learning from Human Feedback (RLHF) has enabled significant advancements within language modeling for powerful, instruction-following models. However, the alignment of these models remains a pressing challenge as the policy tends to overfit the learned ``proxy" reward model past an inflection point of utility as measured by a ``gold" reward model that is more performant -- a phenomenon known as over-optimization. Prior work has mitigated this issue by computing a pessimistic statistic over an ensemble of reward models, which is common in Offline Reinforcement Learning but incredibly costly for language models with high memory requirements, making such approaches infeasible for sufficiently large models. To this end, we propose using a shared encoder but separate linear heads. We find this leads to similar performance as the full ensemble while allowing tremendous savings in memory and time required for training for models of similar size.
\end{abstract}

\section{Introduction} 

Modern language models have been ubiquitous in discussions of general-purpose AI systems that can accomplish myriad tasks across many disciplines and with rapidly increasing capabilities \citet{OpenAI2023GPT4TR, bard2023overview, touvron2023llama, bommasani2022opportunities, wei2022emergent}. However, alongside this increase in capabilities, there has been growing concern around the risks of such systems as they are not entirely interpretable and could be misused to cause substantial harm either maliciously or inadvertently \citep{hendrycks2023overview, wang2023decodingtrust, hendrycks2022unsolved}. 

A salient research question is that of \textit{alignment}; given a set of human values, how do we imbue these principles within the behavior of such systems? \citet{perspectivesonDH}. Suppose we can access a ground truth reward model. One observed phenomenon is that training a reward model from preference feedback eventually reaches an inflection point where increasing the policy performance on the proxy stagnates and ultimately degrades the reward credited by the ground truth model -- even though both are consistent with human labels! \citep{pmlr-v202-gao23h}.

This reward over-optimization problem has been identified as a significant technical issue in scaling up learning from human feedback \citet{ouyang2022, casper2023open, coste2023reward, eisenstein2023helping}. Prior work has noted that because the reward models trained from user feedback are only a proxy for their underlying preferences, opting for a high reward can lead to performance degradation. Recent work has proposed mitigations either through ensembling or transforming the RL objective. Still, these methods are either too computationally expensive or analytically intractable to be used with modern language models given limited compute \citep{coste2023reward, eisenstein2023helping}.

In our work, we propose a simple modification to a common strategy of ensembling: instead of maintaining multiple separate reward models, we opt for a shared backbone with different linear heads with the hypothesis that different initialization and training procedures will generate enough diversity. Our contributions show initial experiments validating the utility of such an approach in RLHF, as it is just as performant as using a full ensemble for mitigating overoptimization while requiring less time for training during reward modeling and less memory and time for PPO.

\section{Related Work}

\textbf{Reinforcement Learning from Human Feedback (RLHF)} RL is a powerful framework for learning diverse and performant policies that can tackle a wide array of tasks \citet{suttonrl, abeelirl, dpbellman, haarnoja2017reinforcement,pomdpdrake}. Of particular recent interest are language models, which have shown impressive base capabilities in terms of instruction following. RLHF is a core component of the tuning processes for many of the currently highest-performing and most widely-used language models, improving their ability to follow instructions and align their responses with human preferences \citet{stiennon2022learning, christiano2023deep}. However, tuning through RLHF can introduce a risk of overfitting to a proxy of true reward since the reward model is learned. Prior work analyzes this phenomenon using a fixed ``gold-standard" reward model in place of humans, training proxy reward models from labels it provides \citep{pmlr-v202-gao23h}. 

\textbf{Ensembles for Overoptimisation} Recent work has shown multiple ways to mitigate this overoptimization issue, but the suggestions only apply during training or require expensive copies of multiple reward models for ensembling without deeper analysis as to why they mitigate this issue \citet{coste2023reward}. Related work further discovered that the ensembles are more effective when pretraining from scratch with separate seeds \citet{eisenstein2023helping}. Separately, other work has investigated uncertainty across ensemble members for RLHF, showing that using the same backbone with different linear heads can lead to significant improvement in calibration but that this weakly correlated with performance for summarization tasks \citet{gleave2022uncertainty}. Concurrent work investigates a similar approach through uncertainty estimation to propose a last layer reward modeling scheme with Bayesian uncertainty estimation \cite{zhang2024overcoming}. Alternatively we use a linear layer for each separate reward function in an ensemble and train the reward model by fine-tuning through the base encoder in the language model. The recent work most similar to our approach investigates using either LoRA for fine-tuning a frozen transformer or similary fine-tuning linear heads with the same transformer backbone for a reward modeling ensemble \cite{zhang2024improving}. Our approach differs in that we forgo using the mean or a least confidence bound estimate due to the observation that the min is better suited for minimizing epistemic uncertainty and is simpler to implement. This allows for utilizing the shared features from supervised fine-tuning while enabling the diversity of reward ensembles in a scalable manner with a simple minimum function.

\section{Background \& Methods}
\subsection{PPO}
Proximal Policy Optimization (PPO) is an algorithm in reinforcement learning (RL) that is particularly adept at ensuring gentle policy updates, crucial for the complex dynamics of language models by optimizing reward with a KL constraint to the original policy.
\begin{equation}
\max_\pi \underset{s, a \sim \pi_{\text{old}}}{\mathbb{E}} \left[ \frac{\pi(a|s)}{\pi_{\text{old}}(a|s)} R(s,a) - \beta \text{KL}(\pi_{\text{old}} || \pi) \right]
\end{equation}

\subsection{Reward Learning}
In standard reward modelings given a prompt completion $x$ and a human pairwise binary preference label $y$, we instantiate the reward model using the backbone feature extractor from a pre-trained or model $\mathcal{F}$. We then initialize a linear reward head $\mathcal{H}: \mathbb{R}^{d} \times \mathbb{R}$ with feature dim $d$ giving a reward function for each linear head.
\begin{equation}
    r_{i}(x) = \mathcal{H}_{i}(\mathcal{F}(x)) 
\end{equation}

\subsubsection{Multi-head Reward Learning}

In our approach, the multi-head reward model is structured upon a shared base neural architecture derived from the pre-trained and supervised fine-tuned language model.\textbf{ \textbf{Everything is fixed } except instead of a singular head we design the model to incorporate \textbf{multiple heads}}. Formally let $\mathcal{F} $ is the feature extractor and $\mathcal{H}_i$ is the $i^{th}$ head linear head, the reward $r_i$ for input $x$ using head $i$ is simply the minimum reward from the linear heads:
\begin{equation}
    \hat{r}(x) = \min_{i} \mathcal{H}_i(\mathcal{F}(x))
\end{equation}

\begin{figure}[h!]
    \includegraphics[width=1.0\textwidth]{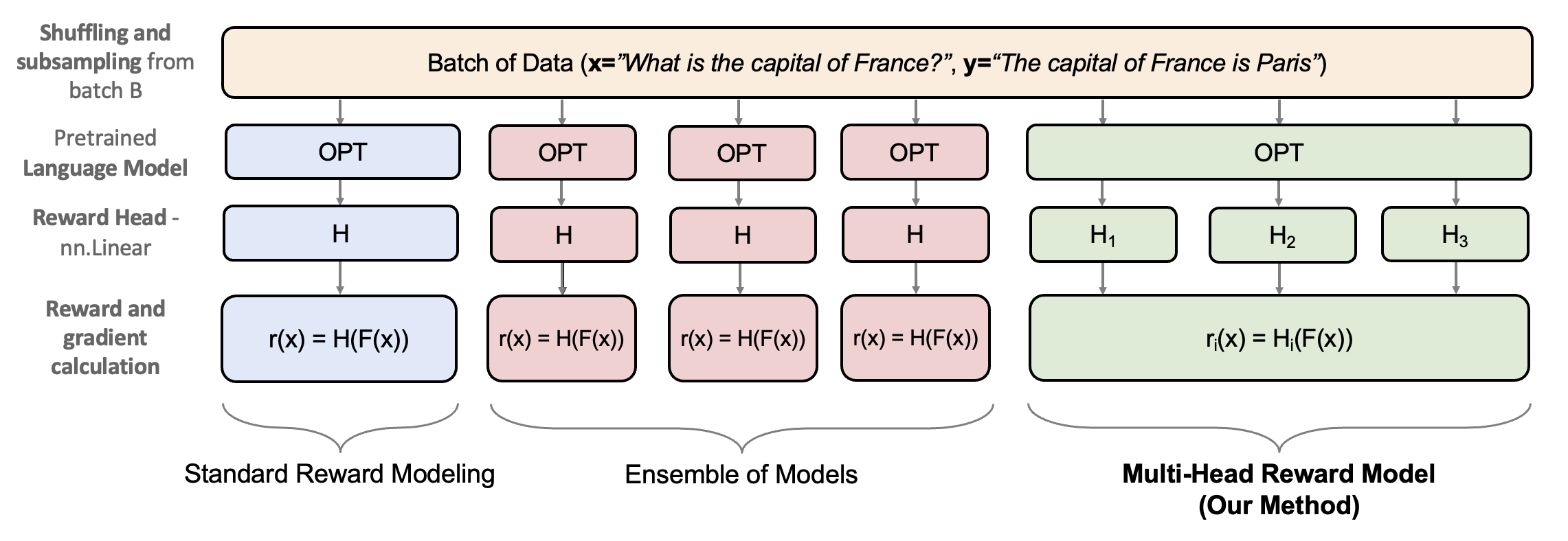}
    \caption{Comparison of Reward Modeling methods to ours (right)\label{fig:model}}
\end{figure}

\section{Experiments}

\subsection{Datasets and Motivation}

We select the \textbf{Alpaca Instructions} dataset, which emphasizes naturalistic, closed-form questions and answers, usually having a well-defined correct response, e.g., the number of planets in the solar system, but also included more open questions such as famous actors who started on broadway, why certain states are named, how to play kickball, etc. \citet{taori2023alpaca}. Given the most relevant prior work studying overoptimisation focused on this dataset, it is our primary dataset for training and evaluation \citep{coste2023reward}.

\subsection{Methodology and Tools}

We follow the RLHF pipeline of supervised fine-tuning (SFT), reward learning, and proximal policy optimization (PPO). The experiments use the OPT model family \citet{zhang2022opt}. We opted for the Alpaca Farm codebase as our framework \citep{dubois2023alpacafarm}.

\begin{itemize}
    \item \textbf{Supervised Fine-Tuning (SFT):} Given a prompt ("Tell me a bedtime story") and some ideal completions, the base model is fine-tuned to minimize perplexity on a split of 52k instructions.
    \item \textbf{Reward Learning:} Given the SFT model and the same set of prompts but now with pairs of completions (preferred and dispreferred), we use the backbone and fine-tune a linear head on the Bradley-terry loss with the preferred completion as the target \citep{Bradley1952RankAO}.
    \item \textbf{PPO} Finally, we further tune the SFT language model as a policy in the RL framework against the reward model through PPO.
\end{itemize}

We modify this pipeline by training the multi-head reward models with the base model initialized from SFT. When selecting the reward for a given sample, since we produce multiple predictions, we are presented with multiple choices for computing our final prediction \citet{coste2023reward}, but we \textbf{simply take the minimum over the ensemble}. A pessimistic estimator might reduce performance in principle. However, for offline RL and RLHF preferences, it has been either conjectured or empirically validated that a minimum is optimal, and prior work demonstrated uncertainty penalties do not improve the efficiency of ensembles in preventing over-optimization over a simple min \citet{zhu2023principled, coste2023reward, kumar2020conservative, chen2021randomized}. Our approach is shown in Figure \ref{fig:model}, and all hyperparameters and further training details are in the Appendix.

\begin{figure}[h!]
    \centering
    \includegraphics[width=0.7\textwidth]{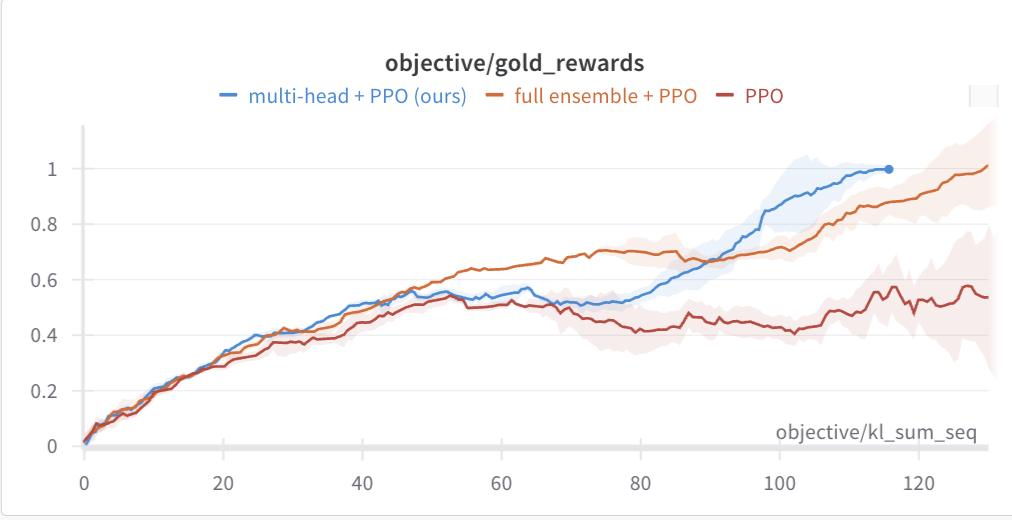}
\end{figure}
\begin{figure}[h!]
    \centering
    \includegraphics[width=0.7\textwidth]{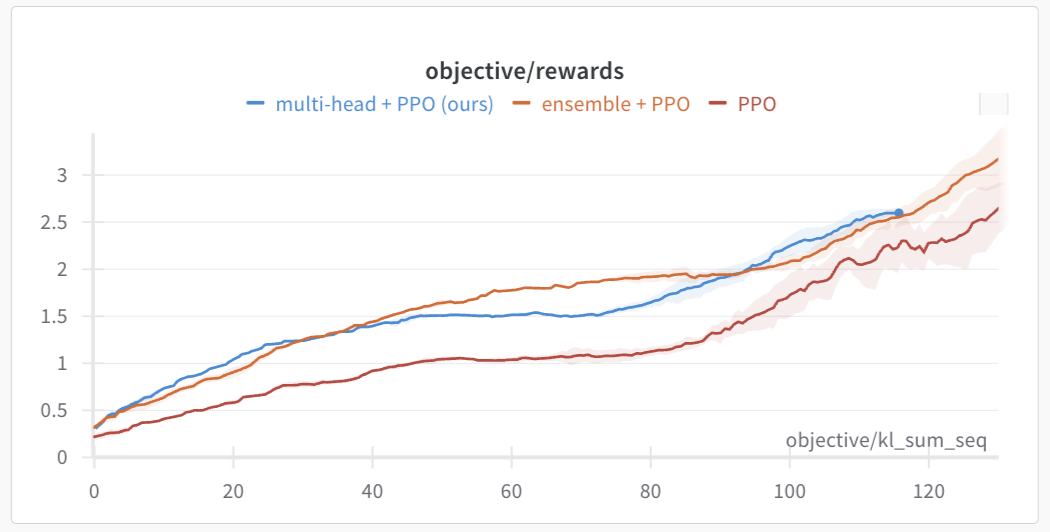}
    \caption{Gold analysis on top, Proxy metrics below.\label{fig:results}}
\end{figure}

\subsection{Results}

To test the efficacy of our approach, we run 1) Standard PPO 2) PPO with an ensemble of three reward models and 3) PPO with a multi-head reward model with three linear heads. We fix ensemble size at three as prior work found no gains when increasing to four or five members due to fixed compute, although we emphasize our method is amenable to a much larger number of ensemble members with minimal overhead \citet{coste2023reward}. We select the 1.3 B parameter OPT model as the proxy and the 6.7B variant as the gold model and run PPO for 15 epochs to extensively test overoptimisation. We also use the 1.3B model as the policy after supervised fine-tuning.

We present our results in Figure \ref{fig:results}.  Given the difference in reward scales and multiple works validating the benefit of using a full ensemble for overoptimisation, we opt for two figures that demonstrate 1) the efficacy of the multi-head ensemble against standard PPO and 2) the efficacy of using a multi-head or full ensemble. Following prior work, we plot reward against KL divergence \citet{pmlr-v202-gao23h, coste2023reward, eisenstein2023helping}. Figure \ref{fig:results} clearly shows the replication of overoptimisation as we reproduce the concave-down nature of gold rewards under standard PPO and then summarily show how using a multi-head reward model bridges this gap. Figure \ref{fig:results} shows that the multi-head and full ensemble nature to similar gold performance however we emphasize that our approach allows an ensemble with much larger reward models and following the prior work suggesting the full ensemble \textbf{we train each member for three epochs} to maximize performance whereas we find the \textbf{multi-head approach needs only one epoch}. We average over three seeds for each method and include further details and ablations in the appendix.

\subsection{Calibration}
\begin{figure}[htbp]
  \centering
  \includegraphics[width=0.6\linewidth]{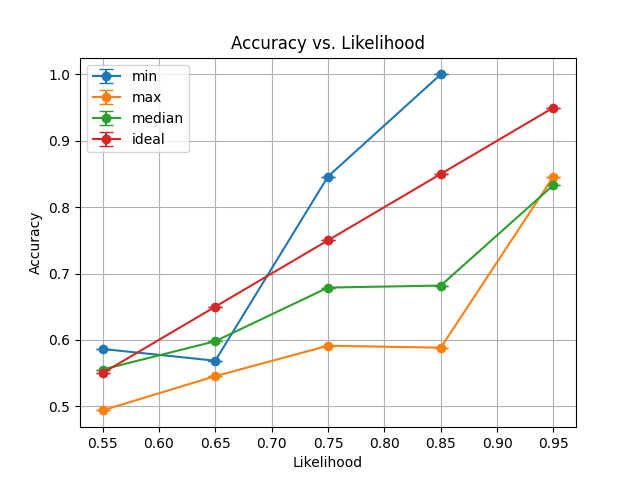}
  \caption{Multi-head model calibration over different objectives with respect to the probabilities (taking min, max over ensemble etc.)}
  \label{fig:yourlabel}
\end{figure}

\subsection{Calibration Analysis and Implications}
We investigate the calibration of our multi-head reward models, as an accurate representation of uncertainty is crucial in RLHF to prevent overoptimistic predictions \cite{casper2023open}. Prior work found that ensembling improves calibration but is weakly related to model error when using a shared backbone with different linear heads, suggesting that separate reward models might be necessary for efficient ensembling \cite{gleave2022uncertainty}. However, we hypothesize that the difference in improvement is due to the diversity of tasks, as the AlpacaFarm dataset focuses on more open-ended questions compared to the summarization tasks in the prior work, which might increase model error due to epistemic uncertainty.
We assess the calibration quality of the models using different ensemble objectives, such as mean and minimum reward estimation. The multi-head models, particularly with the minimum objective, demonstrate an enhanced ability to capture uncertainty in the reward signals without incurring significant computational overhead. While the minimum estimate is technically miscalibrated, it leads to better-than-predicted performance at high levels of certainty. This suggests a regularization effect from the pessimistic estimator, as overoptimized models typically reach a local minima, explaining the "U-shaped" curves that result as KL divergence from the base policy increases.
Our findings align with previous work on pessimistic estimates in ensemble models, underscoring the benefits of a conservative approach in certain RL contexts. Furthermore, our experiments suggest that a smaller number of heads (3) might be optimal for mitigating overoptimisation, providing a valuable guideline for future research in ensemble-based models for RLHF. An increase in the number of heads beyond a certain threshold might introduce more noise than beneficial diversity, potentially leading to overfitting on dataset nuances, which is consistent with existing literature on the effectiveness of smaller ensembles in specific scenarios \cite{galdran2023multihead}. See Figure 3 for details.

\subsection{Conclusion}
Our research contributes a novel method to improve the robustness of aligning language models by utilizing a robust pessimistic statistics while avoiding the cost of materializing a full ensemble. By addressing the challenges of overoptimization computationally efficiently, we move a step closer to developing AI systems that can reliably align with human values and preferences. We plan to further investigate the effects of different datasets, such as the more naturalistic Stanford Human Preferences \cite{shah2019feasibility}
and larger model sizes for future work.

\subsection{Acknowledgements}
AMA acknowledges support from a Knight-Hennessy fellowship and a National Science Foundation fellowship. SK acknowledges support by NSF 2046795 and 2205329, NIFA award 2020-67021-32799, the Alfred P. Sloan Foundation, and Google Inc. 

\bibliography{iclr2024_conference}

\begin{thebibliography}{33}
\providecommand{\natexlab}[1]{#1}
\providecommand{\url}[1]{\texttt{#1}}
\expandafter\ifx\csname urlstyle\endcsname\relax
  \providecommand{\doi}[1]{doi: #1}\else
  \providecommand{\doi}{doi: \begingroup \urlstyle{rm}\Url}\fi

\bibitem[Abbeel \& Ng(2004)Abbeel and Ng]{abeelirl}
Pieter Abbeel and Andrew~Y. Ng.
\newblock Apprenticeship learning via inverse reinforcement learning.
\newblock In \emph{Proceedings of the Twenty-First International Conference on
  Machine Learning}, ICML '04, pp.\ ~1, New York, NY, USA, 2004. Association
  for Computing Machinery.
\newblock ISBN 1581138385.
\newblock \doi{10.1145/1015330.1015430}.
\newblock URL \url{https://doi.org/10.1145/1015330.1015430}.

\bibitem[Bellman(1957)]{dpbellman}
Richard~E Bellman.
\newblock \emph{Dynamic programming.}
\newblock Princeton University Press, Princeton, NJ, USA, 1957.

\bibitem[Bommasani et~al.(2022)Bommasani, Hudson, Adeli, Altman, Arora, von
  Arx, Bernstein, Bohg, Bosselut, Brunskill, Brynjolfsson, Buch, Card,
  Castellon, Chatterji, Chen, Creel, Davis, Demszky, Donahue, Doumbouya,
  Durmus, Ermon, Etchemendy, Ethayarajh, Fei-Fei, Finn, Gale, Gillespie, Goel,
  Goodman, Grossman, Guha, Hashimoto, Henderson, Hewitt, Ho, Hong, Hsu, Huang,
  Icard, Jain, Jurafsky, Kalluri, Karamcheti, Keeling, Khani, Khattab, Koh,
  Krass, Krishna, Kuditipudi, Kumar, Ladhak, Lee, Lee, Leskovec, Levent, Li,
  Li, Ma, Malik, Manning, Mirchandani, Mitchell, Munyikwa, Nair, Narayan,
  Narayanan, Newman, Nie, Niebles, Nilforoshan, Nyarko, Ogut, Orr,
  Papadimitriou, Park, Piech, Portelance, Potts, Raghunathan, Reich, Ren, Rong,
  Roohani, Ruiz, Ryan, Ré, Sadigh, Sagawa, Santhanam, Shih, Srinivasan,
  Tamkin, Taori, Thomas, Tramèr, Wang, Wang, Wu, Wu, Wu, Xie, Yasunaga, You,
  Zaharia, Zhang, Zhang, Zhang, Zhang, Zheng, Zhou, and
  Liang]{bommasani2022opportunities}
Rishi Bommasani, Drew~A. Hudson, Ehsan Adeli, Russ Altman, Simran Arora, Sydney
  von Arx, Michael~S. Bernstein, Jeannette Bohg, Antoine Bosselut, Emma
  Brunskill, Erik Brynjolfsson, Shyamal Buch, Dallas Card, Rodrigo Castellon,
  Niladri Chatterji, Annie Chen, Kathleen Creel, Jared~Quincy Davis, Dora
  Demszky, Chris Donahue, Moussa Doumbouya, Esin Durmus, Stefano Ermon, John
  Etchemendy, Kawin Ethayarajh, Li~Fei-Fei, Chelsea Finn, Trevor Gale, Lauren
  Gillespie, Karan Goel, Noah Goodman, Shelby Grossman, Neel Guha, Tatsunori
  Hashimoto, Peter Henderson, John Hewitt, Daniel~E. Ho, Jenny Hong, Kyle Hsu,
  Jing Huang, Thomas Icard, Saahil Jain, Dan Jurafsky, Pratyusha Kalluri,
  Siddharth Karamcheti, Geoff Keeling, Fereshte Khani, Omar Khattab, Pang~Wei
  Koh, Mark Krass, Ranjay Krishna, Rohith Kuditipudi, Ananya Kumar, Faisal
  Ladhak, Mina Lee, Tony Lee, Jure Leskovec, Isabelle Levent, Xiang~Lisa Li,
  Xuechen Li, Tengyu Ma, Ali Malik, Christopher~D. Manning, Suvir Mirchandani,
  Eric Mitchell, Zanele Munyikwa, Suraj Nair, Avanika Narayan, Deepak
  Narayanan, Ben Newman, Allen Nie, Juan~Carlos Niebles, Hamed Nilforoshan,
  Julian Nyarko, Giray Ogut, Laurel Orr, Isabel Papadimitriou, Joon~Sung Park,
  Chris Piech, Eva Portelance, Christopher Potts, Aditi Raghunathan, Rob Reich,
  Hongyu Ren, Frieda Rong, Yusuf Roohani, Camilo Ruiz, Jack Ryan, Christopher
  Ré, Dorsa Sadigh, Shiori Sagawa, Keshav Santhanam, Andy Shih, Krishnan
  Srinivasan, Alex Tamkin, Rohan Taori, Armin~W. Thomas, Florian Tramèr,
  Rose~E. Wang, William Wang, Bohan Wu, Jiajun Wu, Yuhuai Wu, Sang~Michael Xie,
  Michihiro Yasunaga, Jiaxuan You, Matei Zaharia, Michael Zhang, Tianyi Zhang,
  Xikun Zhang, Yuhui Zhang, Lucia Zheng, Kaitlyn Zhou, and Percy Liang.
\newblock On the opportunities and risks of foundation models, 2022.

\bibitem[Bradley \& Terry(1952)Bradley and Terry]{Bradley1952RankAO}
Ralph~Allan Bradley and Milton~E. Terry.
\newblock Rank analysis of incomplete block designs: I. the method of paired
  comparisons.
\newblock \emph{Biometrika}, 39:\penalty0 324, 1952.
\newblock URL \url{https://api.semanticscholar.org/CorpusID:125209808}.

\bibitem[Casper et~al.(2023)Casper, Davies, Shi, Gilbert, Scheurer, Rando,
  Freedman, Korbak, Lindner, Freire, Wang, Marks, Segerie, Carroll, Peng,
  Christoffersen, Damani, Slocum, Anwar, Siththaranjan, Nadeau, Michaud, Pfau,
  Krasheninnikov, Chen, Langosco, Hase, Bıyık, Dragan, Krueger, Sadigh, and
  Hadfield-Menell]{casper2023open}
Stephen Casper, Xander Davies, Claudia Shi, Thomas~Krendl Gilbert, Jérémy
  Scheurer, Javier Rando, Rachel Freedman, Tomasz Korbak, David Lindner, Pedro
  Freire, Tony Wang, Samuel Marks, Charbel-Raphaël Segerie, Micah Carroll,
  Andi Peng, Phillip Christoffersen, Mehul Damani, Stewart Slocum, Usman Anwar,
  Anand Siththaranjan, Max Nadeau, Eric~J. Michaud, Jacob Pfau, Dmitrii
  Krasheninnikov, Xin Chen, Lauro Langosco, Peter Hase, Erdem Bıyık, Anca
  Dragan, David Krueger, Dorsa Sadigh, and Dylan Hadfield-Menell.
\newblock Open problems and fundamental limitations of reinforcement learning
  from human feedback, 2023.

\bibitem[Chen et~al.(2021)Chen, Wang, Zhou, and Ross]{chen2021randomized}
Xinyue Chen, Che Wang, Zijian Zhou, and Keith Ross.
\newblock Randomized ensembled double q-learning: Learning fast without a
  model, 2021.

\bibitem[Christiano et~al.(2023)Christiano, Leike, Brown, Martic, Legg, and
  Amodei]{christiano2023deep}
Paul Christiano, Jan Leike, Tom~B. Brown, Miljan Martic, Shane Legg, and Dario
  Amodei.
\newblock Deep reinforcement learning from human preferences, 2023.

\bibitem[Coste et~al.(2023)Coste, Anwar, Kirk, and Krueger]{coste2023reward}
Thomas Coste, Usman Anwar, Robert Kirk, and David Krueger.
\newblock Reward model ensembles help mitigate overoptimization, 2023.

\bibitem[Drake(2005)]{pomdpdrake}
Alvin Drake.
\newblock Observation of a markov process through a noisy channel.
\newblock 08 2005.

\bibitem[Dubois et~al.(2023)Dubois, Li, Taori, Zhang, Gulrajani, Ba, Guestrin,
  Liang, and Hashimoto]{dubois2023alpacafarm}
Yann Dubois, Xuechen Li, Rohan Taori, Tianyi Zhang, Ishaan Gulrajani, Jimmy Ba,
  Carlos Guestrin, Percy Liang, and Tatsunori~B. Hashimoto.
\newblock Alpacafarm: A simulation framework for methods that learn from human
  feedback, 2023.

\bibitem[Eisenstein et~al.(2023)Eisenstein, Nagpal, Agarwal, Beirami, D'Amour,
  Dvijotham, Fisch, Heller, Pfohl, Ramachandran, Shaw, and
  Berant]{eisenstein2023helping}
Jacob Eisenstein, Chirag Nagpal, Alekh Agarwal, Ahmad Beirami, Alex D'Amour,
  DJ~Dvijotham, Adam Fisch, Katherine Heller, Stephen Pfohl, Deepak
  Ramachandran, Peter Shaw, and Jonathan Berant.
\newblock Helping or herding? reward model ensembles mitigate but do not
  eliminate reward hacking, 2023.

\bibitem[Galdran et~al.(2023)Galdran, Verjans, Carneiro, and
  Ballester]{galdran2023multihead}
Adrian Galdran, Johan Verjans, Gustavo Carneiro, and Miguel A.~González
  Ballester.
\newblock Multi-head multi-loss model calibration, 2023.

\bibitem[Gao et~al.(2023)Gao, Schulman, and Hilton]{pmlr-v202-gao23h}
Leo Gao, John Schulman, and Jacob Hilton.
\newblock Scaling laws for reward model overoptimization.
\newblock In Andreas Krause, Emma Brunskill, Kyunghyun Cho, Barbara Engelhardt,
  Sivan Sabato, and Jonathan Scarlett (eds.), \emph{Proceedings of the 40th
  International Conference on Machine Learning}, volume 202 of
  \emph{Proceedings of Machine Learning Research}, pp.\  10835--10866. PMLR,
  23--29 Jul 2023.
\newblock URL \url{https://proceedings.mlr.press/v202/gao23h.html}.

\bibitem[Gleave \& Irving(2022)Gleave and Irving]{gleave2022uncertainty}
Adam Gleave and Geoffrey Irving.
\newblock Uncertainty estimation for language reward models, 2022.

\bibitem[Haarnoja et~al.(2017)Haarnoja, Tang, Abbeel, and
  Levine]{haarnoja2017reinforcement}
Tuomas Haarnoja, Haoran Tang, Pieter Abbeel, and Sergey Levine.
\newblock Reinforcement learning with deep energy-based policies, 2017.

\bibitem[Hendrycks et~al.(2022)Hendrycks, Carlini, Schulman, and
  Steinhardt]{hendrycks2022unsolved}
Dan Hendrycks, Nicholas Carlini, John Schulman, and Jacob Steinhardt.
\newblock Unsolved problems in ml safety, 2022.

\bibitem[Hendrycks et~al.(2023)Hendrycks, Mazeika, and
  Woodside]{hendrycks2023overview}
Dan Hendrycks, Mantas Mazeika, and Thomas Woodside.
\newblock An overview of catastrophic ai risks, 2023.

\bibitem[Kumar et~al.(2020)Kumar, Zhou, Tucker, and
  Levine]{kumar2020conservative}
Aviral Kumar, Aurick Zhou, George Tucker, and Sergey Levine.
\newblock Conservative q-learning for offline reinforcement learning, 2020.

\bibitem[Manyika \& Hsiao(2023)Manyika and Hsiao]{bard2023overview}
James Manyika and Sissie Hsiao.
\newblock An overview of bard: An early experiment with generative ai.
\newblock \url{https://ai.google/static/documents/google-about-bard.pdf}, 2023.
\newblock Accessed: 2023-11-26.

\bibitem[OpenAI(2023)]{OpenAI2023GPT4TR}
OpenAI.
\newblock Gpt-4 technical report.
\newblock \emph{ArXiv}, abs/2303.08774, 2023.
\newblock URL \url{https://api.semanticscholar.org/CorpusID:257532815}.

\bibitem[Ouyang et~al.(2022)Ouyang, Wu, Jiang, et~al.]{ouyang2022}
Long Ouyang, Jeffrey Wu, Xu~Jiang, et~al.
\newblock Training language models to follow instructions with human feedback,
  2022.

\bibitem[Russell(2022)]{perspectivesonDH}
Stuart Russell.
\newblock \emph{Artificial Intelligence and the Problem of Control}.
\newblock Springer, 2022.

\bibitem[Shah et~al.(2019)Shah, Gundotra, Abbeel, and
  Dragan]{shah2019feasibility}
Rohin Shah, Noah Gundotra, Pieter Abbeel, and Anca~D. Dragan.
\newblock On the feasibility of learning, rather than assuming, human biases
  for reward inference, 2019.

\bibitem[Stiennon et~al.(2022)Stiennon, Ouyang, Wu, Ziegler, Lowe, Voss,
  Radford, Amodei, and Christiano]{stiennon2022learning}
Nisan Stiennon, Long Ouyang, Jeff Wu, Daniel~M. Ziegler, Ryan Lowe, Chelsea
  Voss, Alec Radford, Dario Amodei, and Paul Christiano.
\newblock Learning to summarize from human feedback, 2022.

\bibitem[Sutton \& Barto(2018)Sutton and Barto]{suttonrl}
Richard~S. Sutton and Andrew~G. Barto.
\newblock \emph{Reinforcement Learning: An Introduction}.
\newblock A Bradford Book, Cambridge, MA, USA, 2018.
\newblock ISBN 0262039249.

\bibitem[Taori et~al.(2023)Taori, Gulrajani, Zhang, Dubois, Li, Guestrin,
  Liang, and Hashimoto]{taori2023alpaca}
Rohan Taori, Ishaan Gulrajani, Tianyi Zhang, Yann Dubois, Xuechen Li, Carlos
  Guestrin, Percy Liang, and Tatsunori~B. Hashimoto.
\newblock Alpaca: A strong, replicable instruction-following model, 03 2023.
\newblock URL \url{https://crfm.stanford.edu/2023/03/13/alpaca.html}.
\newblock Accessed: insert date you accessed here.

\bibitem[Touvron et~al.(2023)Touvron, Lavril, Izacard, Martinet, Lachaux,
  Lacroix, Rozière, Goyal, Hambro, Azhar, Rodriguez, Joulin, Grave, and
  Lample]{touvron2023llama}
Hugo Touvron, Thibaut Lavril, Gautier Izacard, Xavier Martinet, Marie-Anne
  Lachaux, Timothée Lacroix, Baptiste Rozière, Naman Goyal, Eric Hambro,
  Faisal Azhar, Aurelien Rodriguez, Armand Joulin, Edouard Grave, and Guillaume
  Lample.
\newblock Llama: Open and efficient foundation language models, 2023.

\bibitem[Wang et~al.(2023)Wang, Chen, Pei, Xie, Kang, Zhang, Xu, Xiong, Dutta,
  Schaeffer, Truong, Arora, Mazeika, Hendrycks, Lin, Cheng, Koyejo, Song, and
  Li]{wang2023decodingtrust}
Boxin Wang, Weixin Chen, Hengzhi Pei, Chulin Xie, Mintong Kang, Chenhui Zhang,
  Chejian Xu, Zidi Xiong, Ritik Dutta, Rylan Schaeffer, Sang~T. Truong, Simran
  Arora, Mantas Mazeika, Dan Hendrycks, Zinan Lin, Yu~Cheng, Sanmi Koyejo, Dawn
  Song, and Bo~Li.
\newblock Decodingtrust: A comprehensive assessment of trustworthiness in gpt
  models, 2023.

\bibitem[Wei et~al.(2022)Wei, Tay, Bommasani, Raffel, Zoph, Borgeaud, Yogatama,
  Bosma, Zhou, Metzler, Chi, Hashimoto, Vinyals, Liang, Dean, and
  Fedus]{wei2022emergent}
Jason Wei, Yi~Tay, Rishi Bommasani, Colin Raffel, Barret Zoph, Sebastian
  Borgeaud, Dani Yogatama, Maarten Bosma, Denny Zhou, Donald Metzler, Ed~H.
  Chi, Tatsunori Hashimoto, Oriol Vinyals, Percy Liang, Jeff Dean, and William
  Fedus.
\newblock Emergent abilities of large language models, 2022.

\bibitem[Zhang et~al.(2024{\natexlab{a}})Zhang, Chen, Chen, Shen, Sun, and
  Gan]{zhang2024improving}
Shun Zhang, Zhenfang Chen, Sunli Chen, Yikang Shen, Zhiqing Sun, and Chuang
  Gan.
\newblock Improving reinforcement learning from human feedback with efficient
  reward model ensemble, 2024{\natexlab{a}}.

\bibitem[Zhang et~al.(2022)Zhang, Roller, Goyal, Artetxe, Chen, Chen, Dewan,
  Diab, Li, Lin, Mihaylov, Ott, Shleifer, Shuster, Simig, Koura, Sridhar, Wang,
  and Zettlemoyer]{zhang2022opt}
Susan Zhang, Stephen Roller, Naman Goyal, Mikel Artetxe, Moya Chen, Shuohui
  Chen, Christopher Dewan, Mona Diab, Xian Li, Xi~Victoria Lin, Todor Mihaylov,
  Myle Ott, Sam Shleifer, Kurt Shuster, Daniel Simig, Punit~Singh Koura, Anjali
  Sridhar, Tianlu Wang, and Luke Zettlemoyer.
\newblock Opt: Open pre-trained transformer language models, 2022.

\bibitem[Zhang et~al.(2024{\natexlab{b}})Zhang, Ton, Shen, Wang, and
  Liu]{zhang2024overcoming}
Xiaoying Zhang, Jean-Francois Ton, Wei Shen, Hongning Wang, and Yang Liu.
\newblock Overcoming reward overoptimization via adversarial policy
  optimization with lightweight uncertainty estimation, 2024{\natexlab{b}}.

\bibitem[Zhu et~al.(2023)Zhu, Jiao, and Jordan]{zhu2023principled}
Banghua Zhu, Jiantao Jiao, and Michael~I. Jordan.
\newblock Principled reinforcement learning with human feedback from pairwise
  or -wise comparisons.
\newblock \emph{arXiv preprint arXiv:2301.11270}, 2023.
\newblock URL \url{https://arxiv.org/abs/2301.11270}.

\end{thebibliography}
\bibliographystyle{iclr2024_conference}

\appendix
\section{Appendix}
\subsection{D Additional experimental details}

\subsubsection{D.1 Hyperparameters}
We give the hyperparameters here for different components of our RLHF pipeline:
\begin{table}[h]
\centering
\caption{SFT hyperparameters for 6.7B model}
\label{tab:hyperparameters-sft}
\begin{tabular}{|c|c|c|}
\hline
Parameter & Value \\ \hline
Learning rate & 2e-5 \\
Epochs & 3 \\
Batch size & 128 \\ \hline
\end{tabular}
\end{table}

\begin{table}[h]
\centering
\caption{SFT hyperparameters for 1.3B}
\label{tab:hyperparameters-sft}
\begin{tabular}{|c|c|c|}
\hline
Parameter & Value \\ \hline
Learning rate & 8e-6 \\
Epochs & 1 \\
Batch size & 128 \\ \hline
\end{tabular}
\end{table}

\begin{table}[h]
\centering
\caption{RM hyperparameters}
\label{tab:hyperparameters-rm}
\begin{tabular}{|c|c|c|}
\hline
Parameter & Value \\ \hline
Learning rate & 1e-5 \\
Epochs & 3 \\
Batch size & 64 \\ \hline
\end{tabular}
\end{table}

\begin{table}[h]
\centering
\caption{PPO hyperparameters}
\label{tab:hyperparameters-generation}
\begin{tabular}{|c|c|c|}
\hline
Parameter & Value \\ \hline
Max instruction length & 520 \\
Max new tokens (answer length) & 256\\
PPO epochs & 4 \\
Top-p & 0.9 (1.0 for PPO training) \\
Top-k & 0 \\
Temperature & 1.0 \\ 
Rollout Batch size & 512 \\ 
Gradient Step Batch size & 256 \\ 
Learning Rate & 6e-6 \\
\end{tabular}
\end{table}

\subsubsection{D.2 AlpacaFarm Dataset Details}
The AlpacaFarm dataset \cite{dubois2023alpacafarm} employed in our experiments uses the Alpaca data \cite{taori2023alpaca} made up of 52,000 samples. This data is chosen due to its large size and success in training instruction-following models. AlpacaFarm contains five splits: a labeled 10k ”sft” split for supervised fine-tuning, a 10k ”pref” split containing pairwise preference labels, a 20k ”unlabeled” split for training algorithms such as PPO, a 2k validation split, and an unused 10k split. We use the noisy variant of alpaca instructions with 25\% label noise to more closely model real-world data distributions.

\subsubsection{Reward model training ablation}

Interestingly, we find that if we only train each reward model in the full ensemble for one epoch the ensemble approach is not sufficient to prevent over-optimisation, and that at least three epochs are necessary while prior work had five or more. We hypothesize that the multi-head approach is more amenable to underspecified features for reward modeling.

\begin{figure}[h!]
    \centering
    \includegraphics[width=0.7\textwidth]{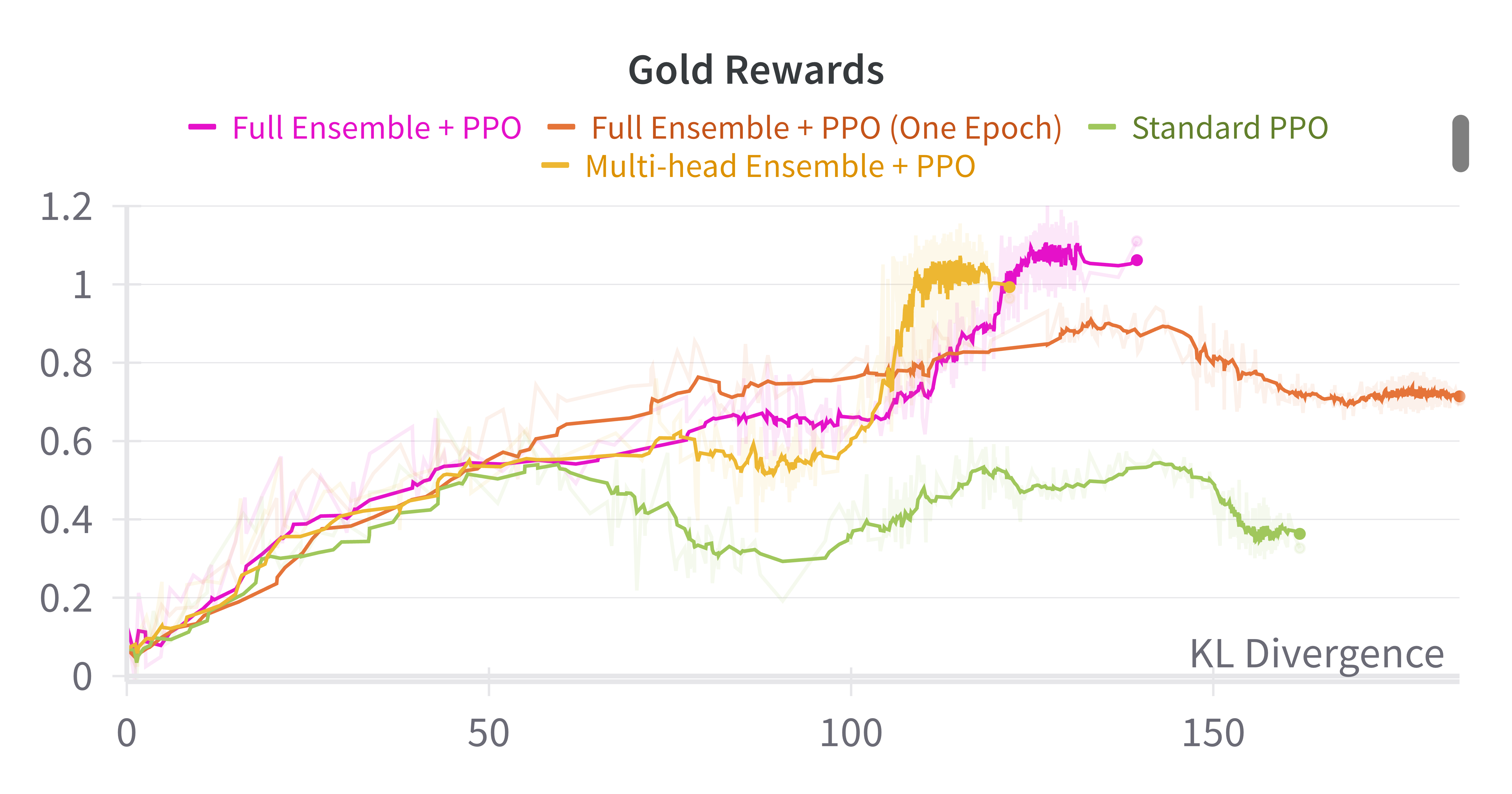}
    \caption{Gold analysis.} 
\end{figure}
\begin{figure}[h!]
    \centering
    \includegraphics[width=0.7\textwidth]{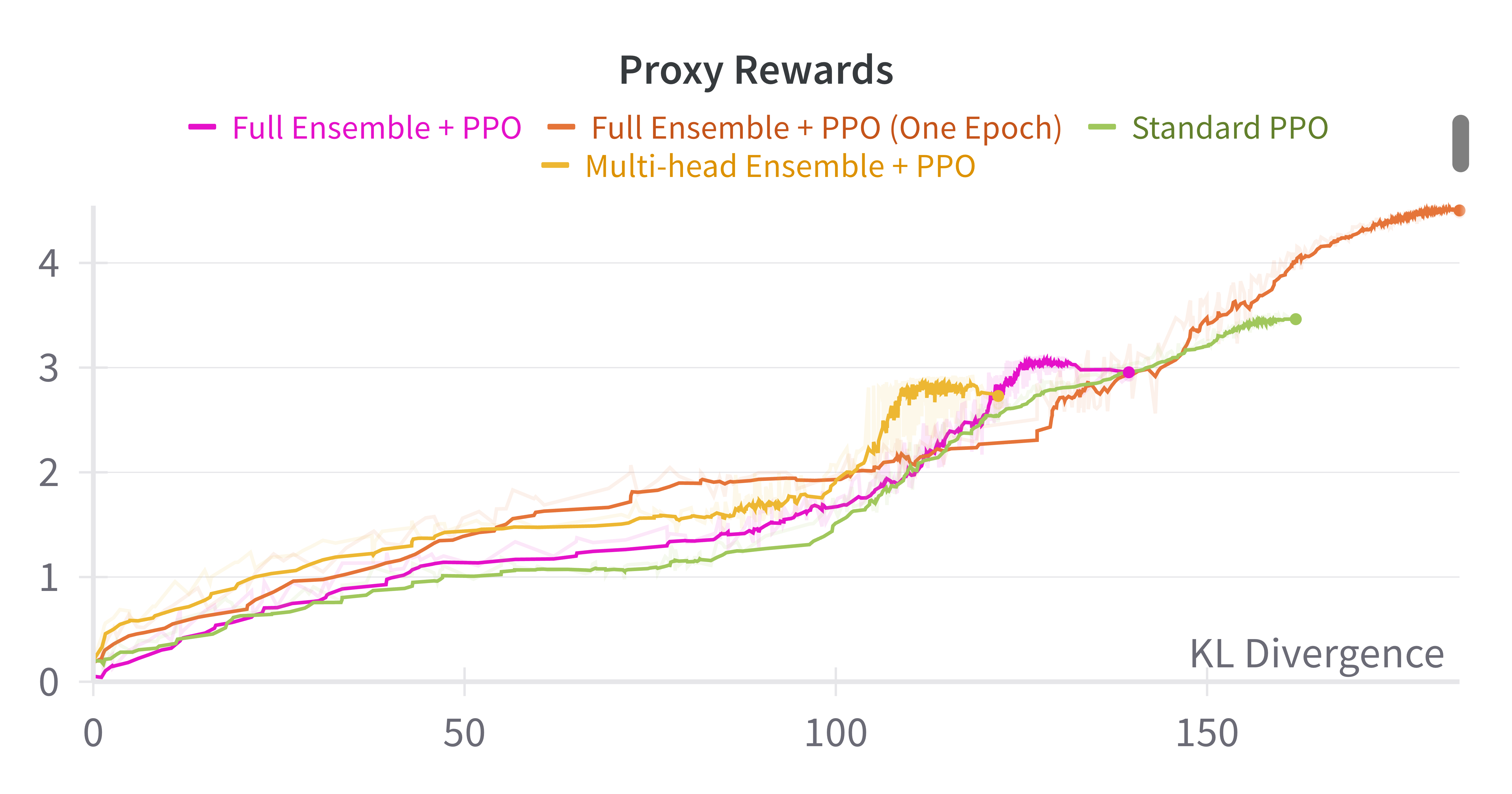}
    \caption{Proxy metrics.} 
    \caption{Gold analysis on top, Proxy metrics below.\label{fig:results}}
\end{figure}

\end{document}